\documentclass{article}
\usepackage{ijcai22}

\usepackage{times}
\usepackage{soul}
\usepackage{url}
\usepackage[hidelinks]{hyperref}
\usepackage[utf8]{inputenc}
\usepackage[small]{caption}
\usepackage{graphicx}
\usepackage{amsmath}
\usepackage{amsthm}
\usepackage{booktabs}
\usepackage{algorithm}
\usepackage{algorithmic}
\usepackage{booktabs}
\usepackage{multirow}
\usepackage{amssymb}
\usepackage{xspace}
\usepackage{enumerate}
\usepackage{enumitem}
\usepackage{caption}
\usepackage{bbm}
\usepackage{comment}
\usepackage{caption}
\usepackage{subcaption}

\usepackage{colortbl}
\definecolor{gray}{gray}{.9}
\usepackage[switch]{lineno}

\makeatletter
\DeclareRobustCommand\bmvaOneDot{\futurelet\@let@token\bmv@onedotaux}
\def\bmv@onedotaux{\ifx\@let@token.\else.\null\fi\xspace}
\makeatother

\def\ie{\emph{ie}\bmvaOneDot}

\def\vs{\emph{vs}\bmvaOneDot}
\renewcommand{\comment}[1]{}
\title{CATrans: Context and Affinity Transformer for Few-Shot Segmentation}

\iftrue
\author{
Shan Zhang $^1$\textsuperscript{\thanks{Interns at the Institute of Deep Learning, Baidu Research}}
\and
Tianyi Wu$^{2,3}$
\and
Sitong Wu$^{2,3}$ 
\and
Guodong Guo$^{2,3}$  \textsuperscript{\thanks{Corresponding author}}
\affiliations
$^1$Australian National University, Canberra, Australia\\
$^2$Institute of Deep Learning, Baidu Research, Beijing, China\\
$^3$National Engineering Laboratory for Deep Learning Technology and Application, Beijing, China\\
\emails
Shan.Zhang@anu.edu.au,
wusitong98@gmail.com,
\{wutianyi01, guoguodong01\}@baidu.com
}
\fi

\begin{document}
% \linenumbers
\maketitle

\begin{abstract}
 Few-shot segmentation (FSS) aims to segment novel categories given scarce annotated support images. The crux of FSS is how to aggregate dense correlations between support and query images for query segmentation while being robust to the large variations in appearance and context. To this end, previous Transformer-based methods explore global consensus either on context similarity or affinity map between support-query pairs. In this work, we effectively integrate the context and affinity information via the proposed novel Context and Affinity Transformer (CATrans) in a hierarchical architecture. Specifically, the Relation-guided Context Transformer (RCT) propagates context information from support to query images conditioned on more informative support features. Based on the observation that a huge feature distinction between support and query pairs brings barriers for context knowledge transfer, the Relation-guided Affinity Transformer (RAT) measures attention-aware affinity as auxiliary information for FSS, in which the self-affinity is responsible for more reliable cross-affinity. We conduct experiments to demonstrate the effectiveness of the proposed model, outperforming the state-of-the-art methods. %and provide extensive ablation studies to validate and analyze components in DRT. 
\end{abstract}

\section{Introduction}

Fully supervised semantic segmentation has made tremendous progress in recent years \cite{demo19,demo15}. 
However, these methods rely heavily on a large amount of pixel-wise annotations which requires intensive manual labor, and are incapable of generalizing to new classes with a handful of annotations. In contrast, humans can recognize a new category even with little guidance.
Inspired by this, Few-shot segmentation (FSS) has recently received a growing interest in the computer vision community \cite{PANet,cycle38,cycle39}.

The goal of FSS is to segment the novel class in the query image, conditioned on the given support set which contains only a few support images and the corresponding ground-truth masks. A fundamental challenge is the large intra-class appearance and geometric variations between support-query pairs, so the key issue is how to effectively reason the relationships of paired samples.

Most FSS methods \cite{cats,cycle,iccv21} follow a learning-to-learn paradigm. Specifically, features are extracted from both query and support images, and then passed through a feature matching procedure to transfer the support mask to the query image. The convolutional neural network (CNN)-based approaches \cite{PANet,cycle36,cycle16} condense the masked object features in the support image into a single or few prototypes. Recently, some approaches introduce Transformer-based architecture to establish pixel-wise matching scores between support-query pairs, containing two main technical routes, \ie, the context and affinity.

\begin{figure}[t]
\centering
%trim={<left> <lower> <right> <upper>}
% \includegraphics[trim=2 2 2 2, clip=true,height=4.0cm]{picture/intro.pdf}
\includegraphics[width=1.0 \linewidth]{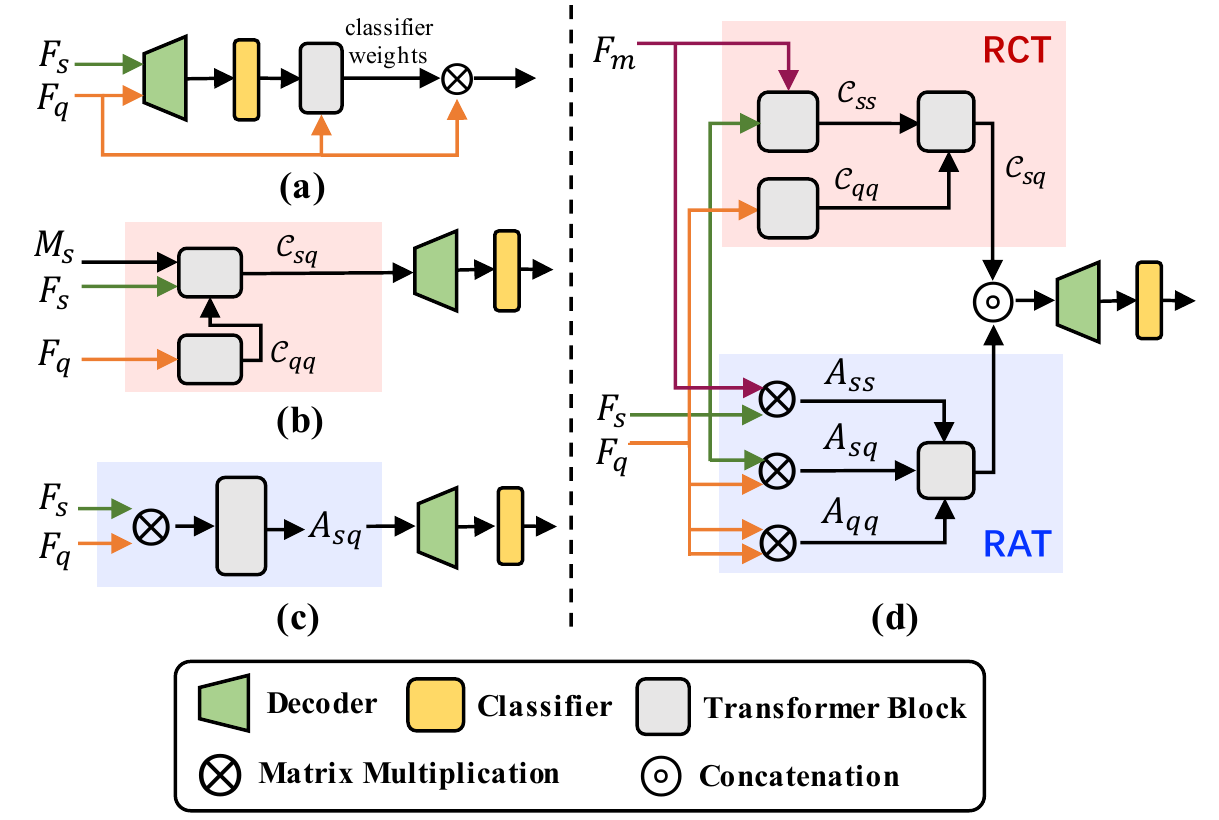}
\caption{Comparisons with different Transformer-based few-shot segmentation methods. The red and blue square shadow denote context ($\mathcal{C}$) and affinity ($\mathcal{A}$) aggregation, respectively. (a) Classifier Weight Transformer \protect\cite{iccv21}. (b) Cycle-Consistent Transformer \protect\cite{cycle}. (c) Cost Aggregation Transformer \protect\cite{cats}. (d) Our Context and Affinity Transformer with Relation-guided Context Transformer (RCT) and Relation-guided Affinity Transformer (RAT).}
\label{intro}
\vspace{-0.4cm}
\end{figure}

% context
% Transformer
%In order to address these shortcomings, recent research introduce Transformer-based architecture to establish long-range \textbf{context} relationships between support-query pairs.
As the dense \textbf{context} information is beneficial to FSS task, especially when the large intra-class variances exist in the support and query set, \cite{iccv21} propose the Classifier Weight Transformer (CWT) to dynamically adapt the classifier's weights trained on support set to each query image as shown in Figure \ref{intro} (a). \cite{cycle} aggregates the context information within query images and between support-query pairs via transformer blocks for self- and cross- alignment,
%the transformer block for cross-alignment to globally consider the varying conditional information from query to support images but with only the query self-alignment by aggregating its relevant context information, 
as shown in Figure \ref{intro} (b). This method, however, suffers from unrepresentative of support feature which stimulates us to propose the Relation-guided Context Transformer (RCT), in which the global context information of support can be considered with a mask encoder \cite{dgp} and a transformer block. The RCT takes as input discriminative self-feature to build more accurate cross-correlation of context, as briefly shown in Figure \ref{intro} (d). 
 
 The attention-aware \textbf{affinity} is globally constructed between support and query features as another guidance for query segmentation. The Cost Aggregation with Transformers (CATs) \cite{cats} builds the cross-affinity between support and query features followed by transformer blocks, as shown in Figure \ref{intro} (c). However, this method does not incorporate individual self-affinity for support object or query image to disambiguate noisy correlations, which measures pixel-wise correspondences within itself, enabling each spatial fiber to match itself and other tokens. %making it robust to the large variations in object appearance between the support and query images. 
 We, thus, design a Relation-guided Affinity Transformer (RAT) for generating a reliable cross-affinity map inherited from the self-affinity. The illustration is schematically depicted in the Figure \ref{intro} (d).
%incorporating self- and cross- affinity together for preserving not only the relative (conditioned on support features) but also absolute correspondences for query segmentation. The thumbnail illustration is presented in Figure \ref{intro} (d).  

Additionally, we explore how to utilize both context and affinity guidance simultaneously. To be specific, we develop a hierarchical CATrans: Context and Affinity Transformer, where we leverage a stack of multi-level correlation maps related to both context and affinity. Moreover, following by \cite{dgp}, we also concatenate the query embedding with high resolution with those low-resolution correlation maps to guide the decoder. 
%that desired interrelation extracted from high-level semantic feature maps should be aligned at discontinuities with precise localization of object boundaries, 

Overall, our contributions are summarized as follows:
\vspace{-0.1cm}
\renewcommand{\labelenumi}{\roman{enumi}.}
\begin{enumerate}[leftmargin=0.6cm]
\item We design a Relation-guided Context Transformer (RCT) with the enhanced support features to propagate informative semantic information from  support to query images.
\item We develop a Relation-guided Affinity Transformer (RAT) to measure the reliable cross correspondences by considering the auxiliary self-affinity of both support object and query images.
\item We propose Context and Affinity Transformer, dubbed as CATrans, in a hierarchical architecture to aggregate the context and affinity together, resulting in discriminative representations from support to query mask, enhancing robustness to intra-class variations between support and query images. Our CATrans outperforms the state-of-the-art methods on two benchmarks, Pascal-5$^i$ and COCO-20$^i$.
\end{enumerate}

\begin{figure*}[t]
\centering
%trim={<left> <lower> <right> <upper>}
\includegraphics[trim=0 0 0 2, clip=true, width=1.0 \linewidth]{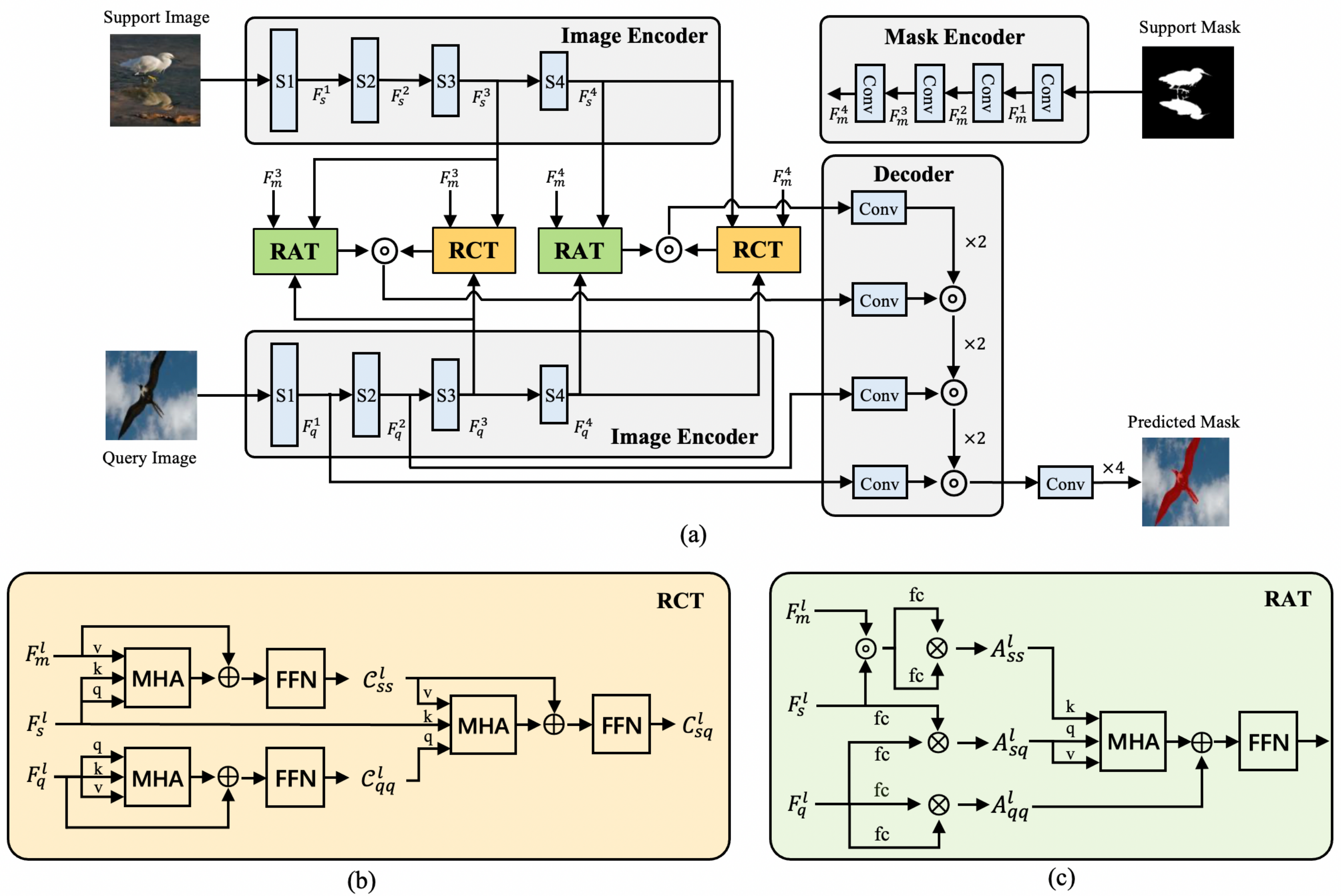}
\caption{
(a) The overall framework of our Context and Affinity Transformer (CATrans). The detailed architecture of our Relation-guided Context Transformer (RCT) and Relation-guided Affinity Transformer (RAT) are shown in (b) and (c), respectively.}
\label{pipeline}
\vspace{-0.1cm}
\end{figure*}

\section{Related work}
\noindent\textbf{Semantic Segmentation.}
Semantic segmentation is a fundamental problem in computer vision, which aims to classify each pixel of an image into predefined categories. Most existing semantic segmentation methods are based on fully convolutional networks (FCNs) \cite{demo19}, that replaces the fully connected layer with fully convolutional ones for pixel-level prediction. Recent breakthroughs in semantic segmentation have mainly come from multi-scale feature aggregation or attention mechanisms. However, the traditional fully supervised segmentation methods require large amounts of image-mask pairs for training, which are very expensive and time consuming. Additionally, it cannot extend model’s generalizability to unseen classes with only a few well-labeled samples.

\noindent\textbf{Few-shot Segmentation.} 
Few-shot semantic segmentation has attracted lots of research attentions after that \cite{trans8}, which first dealt with this issue by proposing to adapt the classifier for each class, conditioned on the support set.  Recent approaches formulate few-shot segmentation from the view of metric learning. %For instance, \cite{PANet,trans9} involves with learning prototypes. 
\cite{trans9} learned prototypes for different classes and the segmentation results are made by cosine similarity between the features and the prototypes. \cite{PANet} developed an efficient prototype learning framework to build consistent prototypes. PFENet \cite{PFENet} made progress by further designing an effective feature pyramid module and leveraged a prior map to achieve a better segmentation performance. Recently, \cite{cycle16,cycle36} found that it is insufficient to represent a category with a single support prototype. Therefore, they used multiple prototypes to represent the support objects via the EM algorithm or K-means clustering. However, these methods disregard the pixel-wise relationships of spatial structure in feature maps. 

Recent works \cite{demo39,cycle,cats,iccv21} attempted to fully utilize a correlation map to leverage the pixel-wise relationships between support and query features. Specially, \cite{demo39} used graph attention networks to propagate information from the support image to query images, and \cite{cycle} utilized cycle-consistent transformer to aggregate the pixel-wise support features into the query. \cite{iccv21} proposed a classifier weight transformer where the transformer is applied to adapt the classifier solely by freezing the encoder and decoder. However, all these dense matching methods focus on either context correspondences or affinity maps only. This is no study about whether these two measures are complementary, and can be integrate to achieve a better performance.

\noindent\textbf{Transformers in Vision.} 
Recently, transformers, first introduced in natural language processing \cite{cycle28}, and are receiving increasing interests  in the computer community. Since the pioneer works such as ViT \cite{trans12}, it demonstrates the pure transformer architecture can achieve the state-of-the-art for image recognition. On the other hand, DETR  \cite{trans14} built up an end-to-end framework with a transformer encoder-decoder on top of backbone networks for object segmentation. And its deformable variants \cite{cycle44} improved the performance and training efficiency. However, there are few studies that compute both context and affinity.

\section{Preliminaries}\label{trans}
\noindent\textbf{Problem Formulation.}
The goal of few-shot segmentation is to segment novel objects with very few annotated samples. Specifically, all classes are divided into two disjointed class set $\mathit{C}_{train}$ and $\mathit{C}_{test}$, where $\mathit{C}_{train} \cap \mathit{C}_{test}=\!\!\emptyset $. To mitigate the overfitting caused by insufficient training data, we follow the common protocol called episodic training. Under K-shot setting, each episode is composed of a support set $\mathcal{S} \!=\! \{(\mathit{I_s, M_s})\}^K$, where $\mathit{I_s}, \mathit{M_s}$  are support image and its corresponding mask, and a query sample $\mathcal{Q} \!=\! (\mathit{I_q, M_q})$, where  $\mathit{I_q}, \mathit{M_q}$  are the query image and mask, respectively. In particular, given dataset $\mathit{D}_{train}=\{\mathcal{S}, \mathcal{Q}\}^{\mathit{N}_{train}}$  and $\mathit{D}_{test}=\{\mathcal{S}, \mathcal{Q}\}^{\mathit{N}_{test}}$ with category set  $\mathit{C}_{train}$ and $\mathit{C}_{test}$,  respectively, where $\mathit{N}_{train}$ and $\mathit{N}_{test}$ is the number of episodes for training and test sets. During training, our model takes a sampled episode from both support masks $\mathit{M_s}$ and query masks $\mathit{M_q}$, and only use support masks for predicting the query segmentation map $\hat{M}_q$ during testing.\\
\noindent\textbf{Revisiting of Transformer.}
Transformer blocks \cite{cycle28} are of a network architecture based on attention mechanisms, which is composed of alternating layers of multi-head attention (MHA) and multi-layer perception (MLP) with inputs of a set of Query (Q), Key (K) and Value (V) elements. %, the MHA adaptively aggregates the value (V) contents according to the attention weights that measure the compatibility of query-key pairs. 
In addition, LayerNorm (LN)  and residual connection are available at the end of each block. Specially, an attention layer is formulated as Architecture Overview:
\vspace{-0.1cm}
\begin{equation}
\text{Atten}(Q,K,V)=softmax\Big(\frac{Q \cdot K^T}{\sqrt{d}}\Big) \cdot V,
\label{eq:atten}
\vspace{-0.1cm}
\end{equation}
where [$Q;K;V$] = [$W_qF_q;W_kF_k;W_vF_v$], in which $F_q$ is the input query sequence, $F_k/F_v$ is the input key/value sequence, $W_q, W_k, W_v \in \mathcal{R}^{c \times c} $ are learnable weights, $c$ is the channel dimension of the input sequences. \\
The multi-head attention layer is an enhancement of attention layer, where $h$ attention units are applied and then concatenated together. Concretely, this operation splits input sequences along the channel dimension $c$ into $h$ groups:
\vspace{-0.01cm}
\begin{equation}
\text{MHA}(Q,K,V)=[head_1,\dots,head_h],
\label{eq:mha}
\vspace{-0.05cm}
\end{equation}
where $head_m \!\!\!\!= \!\!\!\!\text{Atten}(Q_m, K_m, V_m)$ and the inputs [$Q_m,K_m,V_m$] are the $m^{th}$ group from [$Q,K,V$] with dimension $c/h$. %($h\!=\!1$ by default in DRT pipeline, unless otherwise specified).
%shan

\section{Methodology}
Below we present the overall architecture of CATrans followed by a description of its individual components.
%In this section, we first present the overall architecture of our Context and Affinity Transformer (CATrans), followed by a description of the Relation-guided Context Transformer (RCT) and Relation-guided Affinity Transformer (RAT).
\comment{
\subsection{Problem Formulation}
The goal of few-shot segmentation task is to segment novel objects with very few annotated samples. Specifically, all classes are divided into two disjointed class set $\mathit{C}_{train}$ and $\mathit{C}_{test}$, where $\mathit{C}_{train} \cap \mathit{C}_{test}=\!\!\emptyset $. To mitigate the overfitting caused by insufficient training data, we follow the common protocol called episodic training. Under K-shot setting, each episode is composed of a support set $\mathcal{S} \!=\! \{(\mathit{I_s, M_s})\}^K$, where $\mathit{I_s}, \mathit{M_s}$  are support image and its corresponding mask, and a query sample $\mathcal{Q} \!=\! (\mathit{I_q, M_q})$, where  $\mathit{I_q}, \mathit{M_q}$  are a query image and mask, respectively. In particular, given dataset $\mathit{D}_{train}=\{\mathcal{S}, \mathcal{Q}\}^{\mathit{N}_{train}}$ and $\mathit{D}_{test}=\{\mathcal{S}, \mathcal{Q}\}^{\mathit{N}_{test}}$ with category set  $\mathit{C}_{train}$ and $\mathit{C}_{test}$ respectively, where $\mathit{N}_{train}$ and $\mathit{N}_{test}$ is the number of episodes for training and test set. During training, our model takes a sampled episode from both support masks $\mathit{M_s}$ and query masks $\mathit{M_q}$, and only use support masks for predicting the query segmentation map $\hat{M}_q$ during testing.}
\subsection{Architecture Overview}
The overall architecture of our Context and Affinity Transformer (CATrans) is illustrated in Figure \ref{pipeline}, which consists of Image Encoder, Mask Encoder, Relation-guided Context Transformer (RCT), Relation-guided Affinity Transformer (RAT) and the Decoder. Specifically, the input support-query images $\{I_s, I_q\}$ are passed through the Image Encoder to extract multi-scale features $\{F^l_s,F^l_q \!\!\in\!\! \mathbb{R}^{H_l \times W_l \times C_l}\}_{l\!=\!1}^4$, where $l$ denotes the scale-level, and pyramid mask features $\{F^l_m \in \mathbb{R}^{H^l_m \times W^l_m \times C^l_m}\}_{l\!=\!1}^4$ are extracted via the Mask Encoder with the input of support binary mask $M_s$.  The resulting triple $\{F^l_s,F^l_q,F^l_m\}_{l\!=\!3}^4$ is passed into the RCT and RAT, respectively. In practice, the pixel-wise context $\mathcal{C}^l \in \mathbb{R}^{H_l \times W_l \times C_m^l}$ provided by the RCT and dense affinity map $\mathcal{A} \in \mathbb{R}^{H_lW_l \times H_lW_l}$ generated by RAT are concatenated for information aggregation. In the decoder, the fused representations associate with the high-resolution features of query image  $\{F^l_s,F^l_q\}_{l\!=\!1}^2$ for predicting the query mask $\hat{M}_q$. 

%---------------- RCT ----------------
\subsection{Relation-guided Context Transformer}
\textbf{Inspirations:} Current CNN-based prototypical learning approaches, \ie \cite{PANet}, condense the support features into a single or few context-wise prototypes. However, prototypical features inevitably drop spatial information and fail to discriminate incorrect matches due to limited receptive fields of CNNs. So recent research applied the Transformer-based architecture, such as \cite{cycle}, to establish long-range and pixel-wise context relationships between paired support-query samples, outperforming previous CNN-based methods by a large margin. %However, there is a large intra-class differences between support and query images. So 
We conjecture that the representative features within the individual support and query image would help to aggregate more precise cross-relationships, being robust to large intra-class differences of paired support-query samples. %It is illustrated in Figure \ref{RCT} where pink dashed box is the auxiliary module for support self-alignment. 

 To this end, the RCT is designed and shown in Figure \ref{pipeline} (b). We flatten the given triple features $(F^l_s,F^l_q,F^l_m)$ into 1D tokens as inputs for the following items.

\noindent\textbf{Self-context.} Self-context is separately employed for support objects and query features by aggregating its relevant context information, leading to more informative support and query features to be connected via the cross-context. The resulting contexts $\mathcal{C}_{ss}^l$ and $\mathcal{C}_{qq}^l$ are designed as:
\begin{align}
\mathcal{C}_{ss}^l=\text{MLP}(\text{LN}(\text{MHA}(F_s^l,F_s^l,F_m^l))),\\
\mathcal{C}_{qq}^l=\text{MLP}(\text{LN}(\text{MHA}(F_q^l,F_q^l,F_q^l))),
\label{eq:equa1}
\end{align}
where MHA ($\cdot$), LN($\cdot$) and MLP($\cdot$) are operations introduced in the section \ref{trans}.

\noindent\textbf{Cross-context.} Considering that $\mathcal{C}_{ss}^l$, guided by mask features $F_m^l$, mainly focuses on the foreground, but background support pixels are beneficial for building the semantic relationships, we collaborate the enhanced self-context features of support and query with $F_s^l$ to establish the pixel-wise cross-context. The process of cross-context is formed as:
\begin{equation}
\mathcal{C}_{sq}^l=\text{MLP}(\text{LN}(\text{MHA}(\mathcal{C}_{qq}^l,F_s^l,\mathcal{C}_{ss}^l))),
\label{eq:selfalign}
\vspace{-0.1cm}
\end{equation}
where $\mathcal{C}_{sq}^l \in \mathcal{R} ^{H_lW_l \times C_m^l}$ is spatially rearranged to the shape of $H_l \times W_l \times C_m^l$.

%-----------------------------------------------------------
\subsection{Relation-guided Affinity Transformer}
A huge feature distinction between the support
and query images brings barriers for context knowledge transfer, which cripples the segmentation performance. We explore several attention-aware affinity maps that measure pixel-wise correspondences to facilitate the FSS task, as shown in Figure \ref{pipeline} (c). Overall, this module provides affinity guidance stemming from attention-aware features instead of semantics.

\noindent\textbf{Why needs self-affinity?} The training samples belonging to the same class always have features with large variations in appearance as these objects are taken in unconstrained settings. Take aeroplane as an example, all aeroplanes are made by metal and have wings. These features can be seen as intrinsic features. As the differences of shooting angle and lighting conditions, the shape and color of aeroplanes can be different. In few-shot segmentation, we need to enable each pixel-wise feature belonging to itself to match pixel-wise feature at the same position, making it robust to large variations  in  object  appearance  between the support and query images.

\noindent\textbf{What is Self-affinity?} For a support image, we utilize high-dimension support mask features $F_m^l$ concatenated with support image features $F_s^l$ for estimating its affinity map by scaled-dot product followed by softmax function. The process for $l$-th features can be defined as: 
\begin{equation}
\mathcal{A}_{ss}^l=\text{softmax} \Big(\frac{(f^l_m||f^l_s)W_q \cdot ((f^l_m||f^l_s)W_k)^T}{\sqrt{C_l+C_m^l}} \Big),
\label{eq:key}
\vspace{-0.1cm}
\end{equation}
where $\{\cdot||\cdot\}$ denotes the concatenation operation, softmax ($\cdot$) is a row-wise softmax function for attention normalization and two individual fc layers are applied to learn discriminative features by learnable parameters. %$W_q/W_k \in \mathcal{R}^{(C_l+C_m^l) \times (C_l+C_m^l)}$. 
Analogy to support feature, we formulate the  self-affinity for query as:
\begin{equation}
\mathcal{A}_{qq}^l=\text{softmax}\Big(\frac{f^l_qW_q \cdot (f^l_qW_k)^T}{\sqrt{C_l}}\Big).
\label{eq:key2}
\vspace{-0.1cm}
\end{equation}
\noindent\textbf{Cross-affinity.} The cross-affinity between support and query features $F_{ca}^l$ is computed by:
\begin{equation}
\mathcal{A}_{sq}^l=\text{softmax}\Big(\frac{f^l_qW_q \cdot (f^l_\textbf{s}W_k)^T}{\sqrt{C_l}}\Big).
\label{eq:residual}
\end{equation}
However, solely relying on the cross-affinity between features often suffers from the challenges caused by large intra-class variations. We then embed the self-affinity of query $\mathcal{A}_{qq}^l$ into cross-affinity. The final cross-affinity is formulated as:
\begin{equation}
\text{MLP}(\text{LN}(\text{MHA}(\mathcal{A}_{sq}^l,\mathcal{A}_{ss}^l,\mathcal{A}_{sq}^l)+\mathcal{A}_{qq}^l)).
\label{eq:residual}
\end{equation}

%where the $F_{ca}^l\!=\!\text{softmax}(\frac{f^l_qW_q \cdot (f^l_sW_k)^T}{\sqrt{C_l}})$ indicates the cross-affinity maps and $W_q/W_k \in  \mathcal{R}^{ c2 \times c2}$ project the $f_l^q$ to query and key for self-affinity maps of query features.

%----------------------- sota 1----------------------------
\begin{table*}[t]
\caption{Comparison with the state-of-the-art in 1-shot and 5-shot segmentation on PASCAL-5$^i$ dataset using the mIoU (\%) as evaluation metric. Best results in bold.}
\centering
\resizebox{1.\textwidth}{!}{
\begin{tabular}{lc|c|ccccc|ccccc}
\toprule[1pt]
\multirow{2.5}{*}{\textbf{Method}} & \multirow{2.5}{*}{\textbf{Venue}} &\multirow{2.5}{*}{\textbf{Backbone}}&\multicolumn{5}{c|}{\textbf{1-shot}}&\multicolumn{5}{c}{\textbf{5-shot}}\\
\cmidrule{4-13}
~&~&~&$5^0$&$5^1$&$5^2$&$5^3$&Mean&$5^0$&$5^1$&$5^2$&$5^3$&Mean\\
\midrule[1pt]
CANet \cite{cycle38} &CVPR19&\multirow{9}{*}{ResNet-50}&52.5 &65.9& 51.3 &51.9 &55.4& 55.5& 67.8 &51.9 &53.2& 57.1\\
PGNet \cite{demo39} &ICCV19&&56.0& 66.9 &50.6 &50.4 &56.0& 57.7& 68.7 &52.9& 54.6& 58.5\\
RPMMs \cite{cycle36} &ECCV20&&55.2& 66.9& 52.6& 50.7 &56.3& 56.3& 67.3 &54.5& 51.0& 57.3\\
PPNet \cite{cycle16} &ECCV20 &&47.8 &58.8 &53.8& 45.6 &51.5 &58.4& 67.8& 64.9& 56.7& 62.0\\
PFENet \cite{PFENet} &TPAMI20&&61.7& 69.5& 55.4& 56.3& 60.8 &63.1& 70.7 &55.8 &57.9& 61.9\\
CWT \cite{iccv21} &ICCV21&&56.3& 62.0& 59.9 &47.2 &56.4 &61.3& 68.5 &68.5 &56.6& 63.7\\
CyCTR \cite{cycle} &NeurIPS21&&\bf{67.8}& 72.8& 58.0& 58.0& 64.2& 71.1& 73.2& 60.5 &57.5& 65.6\\
DGPNet \cite{dgp} &arXiv21 &&63.5 &71.1&58.2 & 61.2 & 63.5 & 72.4 & 76.9 & 73.2& 71.7 & 73.5 \\
\rowcolor{gray} 
\textbf{CATrans} &Ours & &67.6 &\bf{73.2}&\bf{61.3} & \bf{63.2} & \bf{66.3} & \bf{75.1} & \bf{78.5} &\bf{75.1}& \bf{72.5} &\bf{75.3}\\
\midrule
FWB \cite{fwb} &ICCV19&\multirow{7}{*}{ResNet-101}&51.3 &64.5& 56.7 &52.2& 56.2& 54.9 &67.4 &62.2& 55.3& 59.9\\
DAN \cite{dan} &ECCV20&&54.7 &68.6 &57.8& 51.6& 58.2 &57.9& 69.0 &60.1& 54.9 &60.5\\
PFENet \cite{PFENet} &TPAMI20&&60.5 &69.4 &54.4 &55.9& 60.1 &62.8 &70.4 &54.9 &57.6& 61.4\\
CWT \cite{iccv21} &ICCV21&&56.9& 65.2& 61.2 &48.8 &58.0& 62.6 &70.2& 68.8& 57.2& 64.7\\
CyCTR \cite{cycle} &NeurIPS21&&\bf{69.3} &72.7 &56.5 &58.6 &64.3& 73.5& 74.0 &58.6& 60.2 &66.6\\
DGPNet \cite{dgp} &arXiv21 &&63.9 &71.0 &63.0 & 61.4 & 64.8 & 74.1& 77.4 & 76.7 &73.4& 75.4\\
\rowcolor{gray} 
\textbf{CATrans} &Ours & &67.8&\bf73.2 &\bf64.7 & \bf63.2 &\bf 67.2 & \bf75.2& \bf78.4 & \bf77.7 &\bf74.8& \bf76.5\\
\rowcolor{gray} 
\midrule
\textbf{CATrans} &Ours&Swin-T&\bf68.0&\bf73.5 &\bf64.9 & \bf63.7 &\bf 67.6 & \bf75.9& \bf79.1 & \bf78.3 &\bf75.6& \bf77.3\\
\bottomrule[1pt]
\end{tabular}}
\label{VOC}
\end{table*}

%=====================ablation===============
\begin{table*}[t]
\caption{Ablation studies on the effectiveness of RCT and RAT in (a), multi-level context and affinity utilization in (b), and the head number of attention in (c). Best results are shown in bold.}
%\vspace{-0.2cm}
%\centering
\begin{subfigure}[htp]{0.335\linewidth}{\fontsize{8}{8.4}\selectfont 
\setlength{\tabcolsep}{3.pt}{
\begin{tabular}{ccc|cc|cc}
\toprule
\multirow{2}{*}{RAT}&\multirow{2}{*}{RCT}&\multirow{2}{*}{RCT$^\circ$}&\multicolumn{2}{c|}{PASCAL-5$^i$}&\multicolumn{2}{c}{COCO-20$^i$}\\
~&~&&1-shot&5-shot&1-shot&5-shot\\
\midrule
&~&~&58.3&62.7&33.4&42.6\\
\checkmark&~&~&65.1&70.6&41.3&54.9\\
~&\checkmark&~&66.0&72.4&43.9&56.4\\
~&~&\checkmark&65.3&71.3&42.1&55.3\\
\checkmark&\checkmark&&{\bf66.3}&{\bf75.3}&\bf{46.6}&\bf{58.2}\\
\bottomrule
\end{tabular}
\caption{\label{DRT}}}}
\end{subfigure}
\hspace{0.3cm}
\begin{subfigure}[htp]{0.335\linewidth}{
\fontsize{8}{2}\selectfont 
\setlength{\tabcolsep}{3.3pt}{
\begin{tabular}{cc|cc|cc|cc}
\toprule
%$\substack{\text{\fontsize{8}{4}\selectfont 4}\\\text{\fontsize{5}{8}\selectfont(s=16)}}$
\multicolumn{4}{c|}{\#$l$}&\multicolumn{2}{c|}{PASCAL-5$^i$}&\multicolumn{2}{c}{COCO-20$^i$}\\
\midrule
1&2&3&4&1-shot&5-shot&1-shot&5-shot\\ 
\midrule
~&~&\multicolumn{2}{c|}{\multirow{3}{*}{\checkmark}}&62.4&72.5&42.4&55.8\\
\checkmark&~&&&64.8&73.3&44.8&57.9\\
~&\checkmark&&&65.1&74.2&45.3&58.1\\
\midrule
\multicolumn{2}{c}{\multirow{3}{*}{\checkmark}}&~&\checkmark&64.9&74.1&44.8&56.5\\
&&\checkmark&~&61.2&71.7&40.6&53.3\\
&&\checkmark&\checkmark&{\bf66.3}&{\bf75.3}&\bf{46.6}&\bf{58.2}\\
\bottomrule
\end{tabular}
\caption{\label{multi}}}}
\end{subfigure}
\begin{subfigure}[htp]{0.3\linewidth}{
\fontsize{8}{12}\selectfont 
\setlength{\tabcolsep}{6.5pt}{
\begin{tabular}{c|cc|cc}
\toprule
\multirow{2}{*}{\#$h$}&\multicolumn{2}{c|}{PASCAL-5$^i$}&\multicolumn{2}{c}{COCO-20$^i$}\\
~&1-shot&5-shot&1-shot&5-shot\\
\midrule
1&{\bf66.3}&{\bf75.3}&\bf{46.6}&\bf{58.2}\\
2&65.0&73.9&45.1&56.8\\
3&65.1&73.6&45.4&56.5\\
\bottomrule
\end{tabular}
\caption{\label{h}}}}
\end{subfigure}
\vspace{-0.5cm}
\end{table*}
% ================== exp % ================== 

\begin{table*}[t]
\caption{Comparison with the state-of-the-art in 1-shot and 5-shot segmentation on COCO-20$^i$ dataset using the mIoU (\%) as evaluation metric. Best results in bold. }
\centering
\resizebox{1.0\textwidth}{!}{
\begin{tabular}{lc|c|ccccc|ccccc}
\toprule[1pt]
\multicolumn{2}{c|}{\multirow{2.5}{*}{Method}}&\multirow{2.5}{*}{Backbone}&\multicolumn{5}{c|}{1-shot}&\multicolumn{5}{c}{5-shot}\\
\cmidrule{4-13}
~&~&~&$20^0$&$20^1$&$20^2$&$20^3$&Mean&$20^0$&$20^1$&$20^2$&$20^3$&Mean\\
\midrule[1pt]
PANet \cite{PANet} &ICCV19&\multirow{7}{*}{ResNet-50}&31.5 &22.6& 21.5 &16.2&23.0& 45.9 &29.2& 30.6 &29.6 &33.8\\
RPMMs \cite{cycle36} &ECCV20&&29.5 &36.8& 29.0& 27.0 &30.6& 33.8& 42.0 &33.0 &33.3& 35.5\\
PPNet \cite{cycle16} &ECCV20 &&34.5& 25.4 &24.3& 18.6 &25.7 &48.3 &30.9 &35.7& 30.2& 36.2\\
CWT \cite{iccv21} &ICCV21&&32.2& 36.0& 31.6 &31.6 &32.9 &40.1& 43.8& 39.0& 42.4& 41.3\\
CyCTR \cite{cycle} &NeurIPS21&&38.9 &43.0 &39.6& 39.8 &40.3& 41.1 &48.9 &45.2& 47.0 &45.6\\
DGPNet \cite{dgp} &arXiv21 &&43.6 &47.8 & 44.5 & 44.2& 45.0 & 54.7& 59.1& 56.8 & 54.4 & 56.2\\
\rowcolor{gray} 
\textbf{CATrans} &Ours&&\bf46.5 &\bf49.3 & \bf45.6 & \bf45.1& \bf46.6 &\bf 56.3& \bf60.7& \bf59.2 & \bf56.3 &\bf 58.2\\
\midrule
FWB \cite{fwb} &ICCV19&\multirow{5}{*}{ResNet-101}&19.9 &18.0 &21.0 &28.9 &21.2& 19.1& 21.5 &23.9 &30.1& 23.7\\
PFENet \cite{PFENet} &TPAMI20&&34.3 &33.0 &32.3 &30.1 &32.4& 38.5 &38.6 &38.2& 34.3 &37.4\\
CWT \cite{iccv21} &ICCV21&&30.3& 36.6& 30.5 &32.2 &32.4& 38.5 &46.7 &39.4 &43.2& 42.0\\
DGPNet \cite{dgp} &arXiv21 &&45.1 & 49.5 & 46.6 &45.6 & 46.7 & 56.8& 60.4 & 58.4& 55.9 & 57.9 \\
\rowcolor{gray} 
\textbf{CATrans} &Ours&&\bf47.2 & \bf51.7 & \bf48.6 &\bf47.8 & \bf48.8 & \bf58.5& \bf63.4 & \bf59.6& \bf57.2 &\bf 59.7\\
\midrule
\rowcolor{gray} 
\textbf{CATrans} &Ours&Swin-T&\bf47.9 & \bf52.3 & \bf49.2 &\bf48.0 & \bf49.4 & \bf59.3& \bf64.1 & \bf59.6& \bf57.3 &\bf 60.1\\
\bottomrule[1pt]
\end{tabular}}
\label{COCO}
\end{table*}

\section{Experiments}

In this section, we conduct extensive experiments for our CATrans on two widely-used few-shot segmentation benchmarks, PASCAL-5$^i$ and COCO-20$^i$, to demonstrate the effectiveness of our method. %The datasets and implementation details are described in Section \ref{exp1} and \ref{exp2}. We compare our CATrans with the state-of-the-art in Section \ref{exp3}. Finally, we ablate the effectiveness of the key components of our approach, and give some visualization comparisons in Section \ref{exp4}.

% ===== dataset ===== 
\subsection{Datasets}
\label{exp1}
\textbf{PASCAL-5}$^i$ \cite{trans8} is composed of PASCAL VOC 2012 with additional SBD \cite{SBD} annotations, which contains 20 categories split into 4 folds (15/5 categories as base/novel classes).

\noindent 
\textbf{COCO-20}$^i$ \cite{coco} is created from MS COCO where the 80 object categories are divided into four splits (60/20 categories as base/novel classes).
% ===== Implementation Details ===== 
\subsection{Implementation Details}
\label{exp2}
% % Image Encoder
% \textbf{Image \& Mask Encoder.} 
We conduct all experiments on 1 NVIDIA V100 GPU. 
The models are trained for 20k and 40k iterations on PASCAL-5$^i$ and COCO-20$^i$, respectively, with AdamW as the optimizer. The initial learning rate is set to 5e-5 and decays at 10k iteration with a factor of 0.1. 
During training, we first resize the input images to $384\times384$ and $512\times512$ for PASCAL-5$^i$ and COCO-20$^i$, respectively, and then perform the horizontal flip operation randomly. 
We simply use cross-entropy loss with a weight of 1 and 4 for background and foreground pixels, respectively. The BN layers of image encoder are frozen. 
% Image Encoder
For a fair comparision, we employ the widely-used ResNet-50, ResNet-101 and Swin-Transformer as the image encoder. 
% Mask Encoder
The mask encoder includes four light-weight layers, each of which is composed of $3 \times 3$ convolution, BatchNorm and ReLU.

During evaluation, the results are averaged on the randomly sampled 5k and 20k episodes for each fold and 5 runs with different seeds. we report the mean-IoU (mIoU) under both 1-shot (given a single support example) and 5-shot (given five support examples).

% ===== Compare with SOTA ===== 
\subsection{Comparisons with the state-of-the-art}
\label{exp3}
\noindent\textbf{Results on PASCAL-5$^i$.} 
As shown in Table \ref{VOC}, our CATrans outperforms the previous best DGPNet by +2.8/+1.8\% and +2.4/+1.1\% for 1-shot/5-shot mIoU using ResNet-50 and ResNet-101 as the backbone, repectively.
Rendering the Swin-T as the image encoder, our CATrans further achieves 67.6\% and 77.3\% mIoU for 1-shot and 5-shot segmentation.

\noindent\textbf{Results on COCO-20$^i$.} 
Table \ref{COCO} reports the comparisons on the more challenging COCO-20$^i$ dataset. Compared with the previous best DGPNet, our CATrans surpasses it by +1.6/+2.0\% 1-shot/5-shot mIoU using the ResNet-50. When using ResNet-101, our CATrans is +2.1\% and +1.8\% higher than DGPNet for 1-shot and 5-shot protocals, repectively. Equipped with the Swin-T, our CATrans achieves 49.4/60.1\% 1-shot/5-shot mIoU.

%--------------------------------
\subsection{Ablation Study}
\label{exp4}
Here, we conduct extensive ablation studies with ResNet-50 on PASCAL-5$^i$ and COCO-20$^i$ to analyze the effect of the key components in our CATrans. 
%See supplementary materials\footnote{\url{eee}} for more ablation studies.

\noindent\textbf{Effectiveness of RAT and RCT.}
We ablate our CATrans to observe the effectiveness of RAT and RCT modules, as shown in Table \ref{DRT}. We define the baseline, without RAT and RCT modules, that is simply concatenating the support and query features along the channel mode. The variant with either RAT or RCT boosts baseline by up to +6.8/+7.7 on 1-shot protocol. Moreover, it can be seen that the support self-context branch of RCT provides additional $\sim$1\% mIoU on 1-shot setting (RCT \vs RCT$^\circ$), endowing RCT with a powerful context aggregator. Considering that the cross-affinity between support and query images can serve as the additional guidance for FSS task, when the large feature distinctions impede context knowledge transfer, we verify how much benefit comes from the RAT. Table \ref{DRT} shows the model equips with both RCT and RAT has $\sim$5\% increase for FSS results.
%------------------- visual --------------------
\begin{figure}[t]
\begin{minipage}[t]{0.09\textwidth}
\centering
\includegraphics[width=1.7cm]{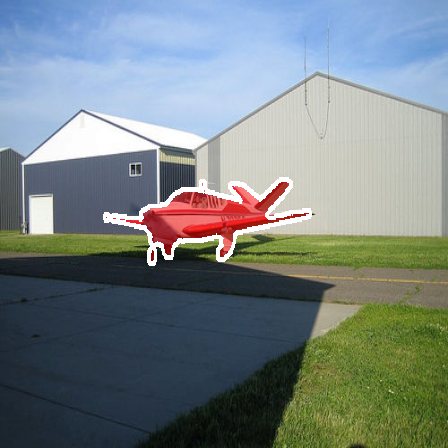}\\
\includegraphics[width=1.7cm]{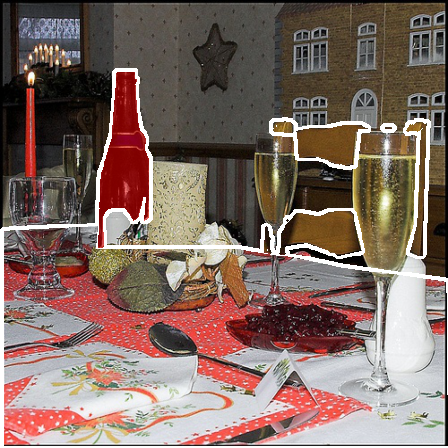}
\caption*{Support set}
\end{minipage}
\hspace{0.0002mm}
\begin{minipage}[t]{0.09\textwidth}
\includegraphics[width=1.7cm]{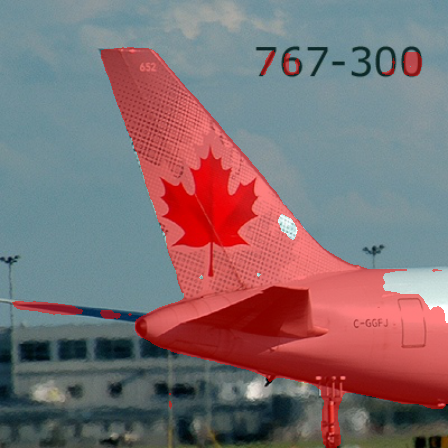}\\
\includegraphics[width=1.7cm]{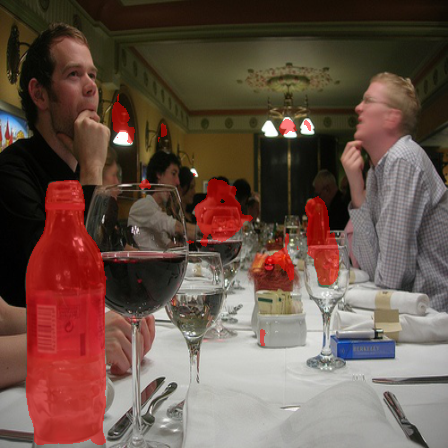}
\caption*{RAT}
\hspace{5mm}
\end{minipage}
\hspace{0.0002mm}
\begin{minipage}[t]{0.09\textwidth}
\includegraphics[width=1.7cm]{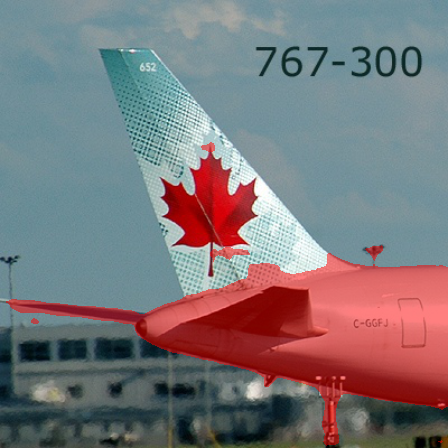}
\includegraphics[width=1.7cm]{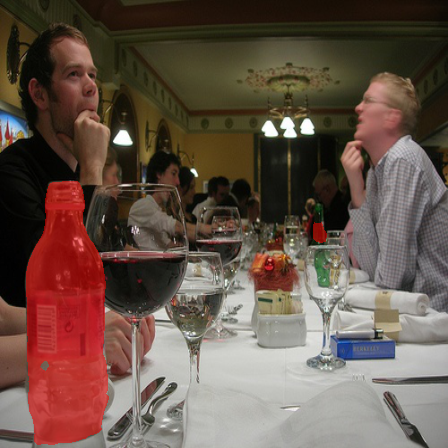}
\caption*{RCT}
\end{minipage}
\hspace{0.0002mm}
\begin{minipage}[t]{0.09\textwidth}
\includegraphics[width=1.7cm]{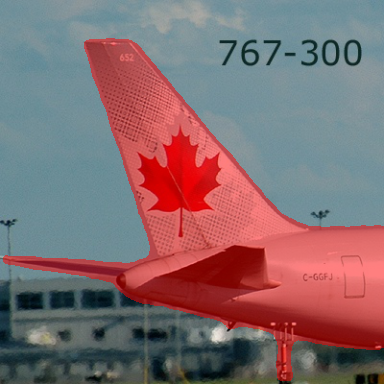}
\includegraphics[width=1.7cm]{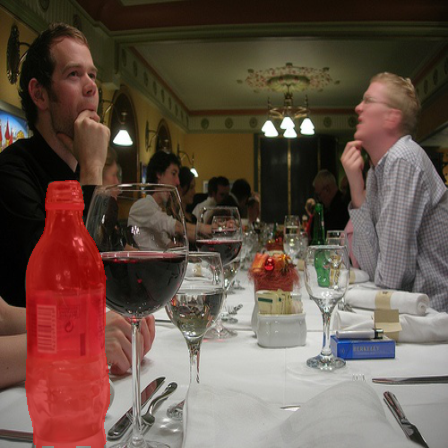}
\caption*{CATrans}
\end{minipage}
\hspace{0.0002mm}
\begin{minipage}[t]{0.09\textwidth}
\includegraphics[width=1.7cm]{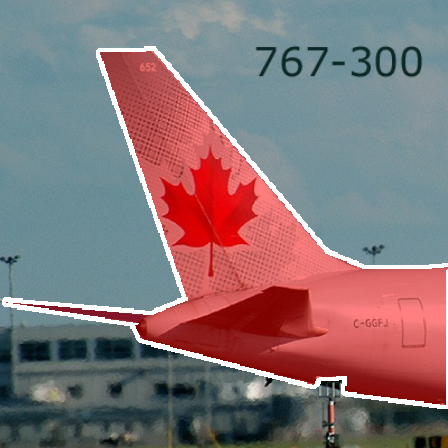}
\includegraphics[width=1.7cm]{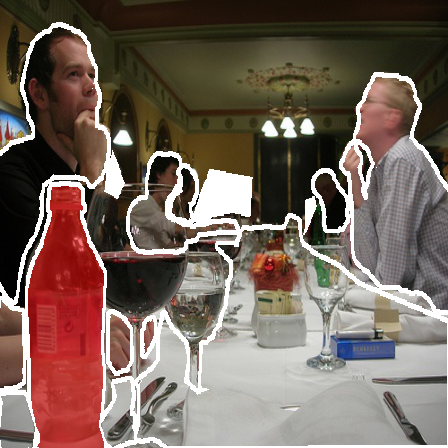}
\caption*{GT}
\end{minipage}
\vspace{-0.4cm}
\caption{Visualize the results predicted by the CATrans and its variants on PASCAL-5$^i$, 1-shot setting. GT means Ground Truth of the query image.}
\vspace{-0.3cm}
\label{cfig}
\end{figure}

\noindent\textbf{Multi-scale Representations.} We first verify the influence of fusing the high-resolution features of query
image ($l\!=\!1/2$) on the top panel of Table \ref{multi}. Then on the bottom panel of Table \ref{multi}, it shows a comparison experiment between single-scale and multi-scale representations of CATrans ($l\!=\!3/4$). The more levels of information used, the better the performance. The performance reaches the highest when these two levels are both leveraged, especially in the 1-shot setting where multi-scale guidance can extract more guiding information for query segmentation with 1.4\%/1.8\% mIoU gain on PASCAL-5$^i$/COCO-20$^i$.

\noindent\textbf{Effect of Model Capacity.}
We stack more numbers of heads of attention layer to increase capacity of  our CATrans and validate its effectiveness. It shows that our model performance is stable across different choices, particularly for $\#h\geq$2.

\noindent\textbf{Memory and Run-time.}
CATrans, with  trivial computational overhead, performs the best over closely related transformer-based methods. CyCTR  and CATs stack two successive transformer blocks whereas CATrans consists of RCT and RAT, each with one transformer block. The memory and run-time comparisons are below: %1.78GB/31.7ms (CyCTR) \vs 1.85GB/33.2ms (ours) \vs 1.90GB/34.5ms (CATs). 
{
\fontsize{8}{2}\selectfont  
\setlength{\tabcolsep}{6pt}{
\hspace{-0.3cm}
  \resizebox{0.45\textwidth}{!}{
      \renewcommand{\arraystretch}{0.5}
\begin{tabular}{ccc|ccc}
\toprule
\multicolumn{3}{c|}{Memory (GB)}&\multicolumn{3}{c}{Run-time (ms)}\\
\midrule
CyCTR&CATrans&CATs&CyCTR&CATrans&CATs\\
\midrule
1.78&1.85& 1.90&31.7& 33.2& 34.5\\ 
\bottomrule
\end{tabular}}
}}

\noindent\textbf{Qualitative results.}
To show the performance of CATrans intuitively, we visualize some final prediction masks produced by our method and its variants in Figure \ref{cfig}. The first column is support image and its ground truth, and the next three columns are query mask produced by RAT, RCT and CATrans, respectively. The last column is ground truth of query image. On the top row of Figure \ref{cfig}, as the large intra-class appearance variations between support and query images hidden context knowledge transferring, RCT fails to precisely segment one of its aerofoils, where RCT performs better by use of the self- and cross- affinity  for query segmentation. On the bottom of Figure \ref{cfig}, the RAT mistakenly regards some plates as part of the target cup because of the similarity in the shape and color, measured by the attention-aware affinity. But RAT reduces the area of wrong segmentation significantly by successfully aggregating the context information. Overall, adopting both RAT and RCT performs the best.
\section{Conclusion}
We have proposed the novel Context and Affinity Transformer (CATrans) with RCT and RAT in a hierarchical architecture to deal with the large intra-class appearance and geometric variations in few-shot segmentation. Different from previous approaches that either build on context or affinity between support and query image, our CATrans effectively incorporate both measures for query segmentation. In addition, we consider pixel-wise correspondences for individual support and query features to disambiguate noisy correlations. %Extensive experiments have shown that our CATrans outperforms the state-of-the-art methods by a significant margin. 
%Moreover, we have conducted extensive ablation studies to validate our choices and explore its capacity.
\bibliographystyle{named}
\bibliography{bibijcai22}

\end{document}